\documentclass[a4paper,oneside,11pt]{article}

\usepackage[american]{babel}
\usepackage[autostyle]{csquotes}

\usepackage{epsfig}
\usepackage{subcaption}
\usepackage{calc}
\usepackage{amssymb}
\usepackage{amstext}
\usepackage{amsmath}
\usepackage{amsthm}
\usepackage{multicol}
\usepackage{pslatex}
\usepackage{algorithm2e}
\usepackage[bottom]{footmisc}
\usepackage{setspace}
\usepackage{csvsimple}
\usepackage{authblk}

\usepackage{natbib}

\usepackage{xcolor}

\usepackage[acronym, nomain, section]{glossaries}
\setacronymstyle{long-short}

\usepackage{tabularx}
\usepackage{booktabs}
\usepackage{microtype}
\usepackage{titlesec}
\titlespacing*{\paragraph}
{0pt}   
{6pt}   
{1em}   
\usepackage{needspace}

\usepackage[hidelinks]{hyperref}
\usepackage{cleveref}
\usepackage{orcidlink}

\newcommand{\cfcite}[1]{\citep[cf.][]{#1}}

\newcommand{\eg}{e.\,g.\@\xspace}
\newcommand{\ie}{i.\,e.\@\xspace}

\renewcommand{\Pr}{\text{Pr}} 
\renewcommand{\Re}{\text{Re}} 
\newcommand{\FOne}{$F_1$\@\xspace} 
\newcommand{\TP}{\text{TP}}
\newcommand{\FP}{\text{FP}}
\newcommand{\FN}{\text{FN}}

\providecommand{\keywords}[1]{%
    \vspace{0.5em}
    {\footnotesize
    \noindent\textbf{Keywords: } #1
    }
}

\newacronym{CRF}{CRF}{Conditional Random Field}
\newacronym{CVET}{CVET}{Continuing Vocational Education and Training}

\newacronym{DAPT}{DAPT}{Domain-adaptive Pre-training}

\newacronym{DNN}{DNN}{Deep Neural Network}

\newacronym{HMM}{HMM}{Hidden Markov Model}

\newacronym{IE}{IE}{Information Extraction}

\newacronym{LLM}{LLM}{Large Language Model}
\newacronym{LSTM}{LSTM}{Long Short-Term Memory}

\newacronym{NAT}{NAT}{Noise-aware Training}
\newacronym{NER}{NER}{Named Entity Recognition}
\newacronym{NLP}{NLP}{Natural Language Processing}

\newacronym{OCR}{OCR}{Optical Character Recognition}

\newacronym{VET}{VET}{Vocational Education and Training}
\newacronym{VLM}{VLM}{Vision Language Model}

\graphicspath{
    {figures/}
}

\title{Noise-Aware Named Entity Recognition for Historical VET Documents\thanks{This is an extended, non-peer-reviewed version of the paper presented at VISAPP 2026.}}

\author[1]{Alexander M. Esser\,\orcidlink{0000-0002-5974-2637}}
\author[1,2]{Jens Dörpinghaus\,\orcidlink{0000-0003-0245-7752}}

\affil[1]{Federal Institute for Vocational Education and Training (BIBB), Bonn, Germany}
\affil[2]{Department of Computer Science, University of Koblenz, Germany}

\affil[ ]{\texttt{alexander.esser@bibb.de, doerpinghaus@uni-koblenz.de}}

\date{}

\begin{document}

\maketitle


\abstract{
    This paper addresses Named Entity Recognition (NER) in the domain of Vocational Education and Training (VET), focusing on historical, digitized documents that suffer from OCR-induced noise.
    We propose a robust NER approach leveraging Noise-Aware Training (NAT) with synthetically injected OCR errors, transfer learning, and multi-stage fine-tuning.
    Three complementary strategies, training on noisy, clean, and artificial data, are systematically compared.
    Our method is one of the first to recognize multiple entity types in VET documents. It is applied to German documents but transferable to arbitrary languages.
    Experimental results demonstrate that domain-specific and noise-aware fine-tuning substantially increases robustness and accuracy under noisy conditions.
    We provide publicly available code for reproducible noise-aware NER in domain-specific contexts.
}




\section{\uppercase{Introduction}}
\label{sec:introduction}

\glsreset{OCR}
\glsreset{NER}
\glsreset{VET}
In this paper, we propose a robust approach to \gls{NER} in the field of labor market research, with a particular focus on \gls{VET}. This field contains both contemporary and historical data; our focus will be on historical, digitized documents. We apply \gls{NAT} to address the challenge of noise induced by \gls{OCR}.


\subsection{Motivation}
\label{sec:motivation}

\glsreset{VET} 
\glsreset{CVET} 
The labor market is constantly evolving, shaped by global developments and country-specific conditions.
In German-speaking countries, technical innovations, such as digitalization, and societal changes require employees to acquire new skills. Specifically, \gls{VET}, retraining, and \gls{CVET} are key to meeting these demands \cfcite{dobischat2019digitalisierung,helmrich2016digitalisierung}. Policymakers in Germany have increasingly recognized the central importance of continuing vocational training and lifelong learning for the success of the digital transformation \cfcite{schiersmann2022weiterbildungsberatung}.

The system offers \gls{VET} as part of tertiary education for professionals~(\emph{Ausbildung}, \emph{Umschulung}).
Here, the term VET is intended to also include CVET, covering advanced and upgrading training (\emph{Weiterbildung}, \emph{Fortbildung}). 

The Federal Institute for Vocational Education and Training (BIBB) holds large datasets related to \gls{VET}, including digitized archival documents.
\citet{Reiser2024} published a dataset of historical \gls{VET} documents dating from 1908 to the present, comprising 2\,125 documents.

\paragraph{Classification Systems and Taxonomies}~\\
The \emph{International Standard Classification of Occupations} (ISCO) was developed by the International Labour Organization (ILO) and first published in 1958\footnote{See\,{}\url{https://www.ilo.org/public/english/bureau/stat/isco/isco08/}.}. Further editions followed in 1968 and 1988, with the most recent version being published in 2008. 
It has also been adopted by the European Union (EU), and specific versions have been developed for some German-speaking countries, including Germany, Austria, and Switzerland. ISCO occupations are structured by skill level and linked to the \emph{European Skills, Competences, Qualifications and Occupations} (ESCO) ontology, adding another hierarchy level to the data~\cfcite{Reiser2024, Dorau2025}.
\needspace{\baselineskip}

\pagebreak[2]

In Germany, the reference classification for the Federal Employment Agency (BA) and its research institute (IAB) is the \emph{German Classification System of Occupations} (Klassifikation der Berufe; KldB). Here, occupations are structured at task level. The latest version is the 2020 revision of KldB 2010, which has been completely redesigned, rendering the previous versions from 1988 and 1992 obsolete. 
It has been developed to be compatible with ISCO-08. The \emph{German Labor Market Ontology} (GLMO) offers all classifications used in Germany~\cfcite{Fischer2024,dorpinghaus2023towards}.

\paragraph{Research Interests in VET}~\\
The automated processing and integration of diverse labor market data sources is considered an important, but complex, endeavor \cfcite{Fischer2024}.
Our long-term goal is to process and structure \gls{VET} datasets to make them available in a data warehouse or as knowledge graphs.
This enables researchers to explore a wide range of related research questions in \gls{VET} and the social sciences more broadly, as well as in fields such as psychology and economics.

Based on the entities recognized in this work, references between \gls{VET} documents can be established and represented in a knowledge graph. Building on this,  further research questions in the field of \gls{VET} can be investigated, \eg, comparisons of historical training regulations in West and East Germany, the accreditation of professional qualifications, or the evolution of job profiles over time.

\paragraph{Document Processing Pipeline}~\\
To achieve this, a document processing pipeline is required to automatically process large volumes of data and extract information.
Following classical architectures~\cfcite{Konya2013}, such a pipeline typically consists of various preprocessing steps, a central \gls{OCR} engine, and subsequent \gls{IE} tasks, such as \gls{NER}.

A major challenge is that \gls{OCR} errors and domain-specific terminology can significantly affect downstream tasks like \gls{NER}. Therefore, we propose a \gls{NAT} approach to increase the robustness of \gls{NER} against \gls{OCR} errors.

\begin{figure}[!ht]
    \centering
    \includegraphics[width=0.7\textwidth]{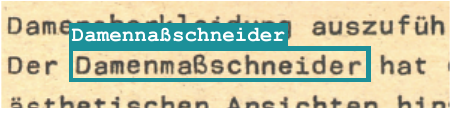}
    \caption{Example of an OCR error in a job title.}
    \label{fig:example_ocr}
\end{figure}

\Cref{fig:example_ocr} provides an illustrative example of the core objective of this work: On a document page of a historical \gls{VET} document, various entities should be recognized. The term \enquote{Damennaßschneider} (\emph{ladies' tailor}) was misspelled during \gls{OCR}; nevertheless, it should be recognized as a \emph{job title} entity during \gls{NER}.


\subsection{Contributions}

The contributions of this paper are:
\begin{itemize}
    \item We propose a robust, noise-aware \gls{NER} approach for \gls{VET} documents.
    \item We compare \emph{domain-specific fine-tuning} and \emph{noise-aware training} strategies by training three models: \emph{noisy}, \emph{clean}, and \emph{artificial}.
    \item We apply \emph{transfer learning} for domain adaptation and \emph{multi-stage fine-tuning} to integrate taxonomies as additional training data.
    \item We present one of the first approaches to recognize multiple types of entities in \gls{VET} data, whereas prior work typically focused on a single entity type.
    \item We subsequently provide a qualitative analysis of recognition rates, confusions, and intrinsic difficulty across different entity types.
    \item We conduct a qualitative error analysis for typical OCR errors in historical \gls{VET} documents.
    \item To support the reproducibility of scientific results, we make our code for training and evaluation publicly available\footnote{Code is available at \url{https://github.com/TM4VETR/noise_aware_ner_vet/}.}.
\end{itemize}


\subsection{Outline}
The remaining paper is structured as follows: 
\Cref{sec:background} provides background information on the document processing pipeline and the entities to be recognized.
\Cref{sec:related_work} discusses related work in the fields of \gls{NER}, \gls{NAT}, and \gls{VET}.
\Cref{sec:method} presents our approach, including details on training data, error injection, and multi-stage fine-tuning.
\Cref{sec:evaluation} evaluates the three complementary model variants \emph{noisy}, \emph{clean}, and \emph{artificial}, as well as the accuracy per entity type.
\Cref{sec:conclusion} concludes the paper and outlines future work.


\section{\uppercase{Background}}
\label{sec:background}

This section provides an overview of the document processing pipeline and the types of entities considered in this study.


\subsection{Document Processing Pipeline}
\label{sec:background-pipeline}

\begin{figure}[!ht]
    \centering
    \includegraphics[width=0.6\textwidth]{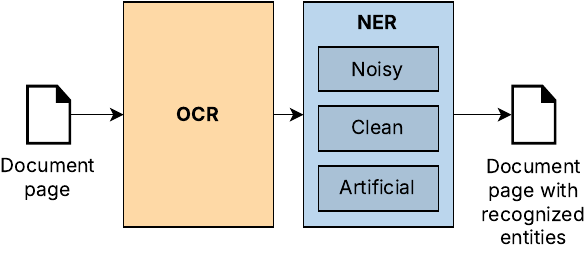}
    \caption{Document processing pipeline.}
    \label{fig:pipeline}
    \vspace{1em}
\end{figure}

This work is part of our broader research agenda. The long-term objective is to build a document processing pipeline for \gls{VET} documents.
In its first version, this pipeline includes modules for \gls{OCR} and \gls{NER}.
\Cref{fig:pipeline} shows the pipeline.

\pagebreak[2]

There are generally two types of documents: \emph{born-digital} documents (mostly in PDF format) and \emph{scanned} documents (in PDF format or various image formats). Many approaches are only capable of handling born-digital PDF documents; in such cases, no quality issues or \gls{OCR} errors occur. For processing historical \gls{VET} data, however, it is crucial to also process scanned documents, since these represent the major share of the data.

Given a document page, \ie, a scanned page image, first \gls{OCR} is applied, using the \emph{Tesseract} OCR engine~\citep{Smith2007}.
Subsequently, \gls{NER} is applied. In this step, three different models, \emph{noisy}, \emph{clean}, and \emph{artificial}, are used, which are described in detail in \Cref{sec:models}. This yields a list of recognized entities for each document page.


\subsection{Named Entities}
\label{sec:background-entities}

During entity recognition, the following named entities are to be identified in German texts:

\begin{enumerate}
    \item Job Titles (\textsc{JOB\_TITLE})
    \item Groups of Job Titles (\textsc{JOB\_TITLE\_GROUP})
    \item Skills (\textsc{SKILL})
    \item Training Subjects (\textsc{SUBJECT})
    \item Work Activities (\textsc{ACTIVITY})
\end{enumerate}

\paragraph{Job Titles (\textsc{JOB\_TITLE})}~\\
As entity \textsc{JOB\_TITLE}, specific occupational titles are to be recognized.
The most frequent job title in the annotated data is \enquote{Facharbeiter/Facharbeiterin} (\emph{skilled worker}).

The task of entity recognition in this paper is \emph{not} to classify a given job title according to a classification system but to identify all \textsc{JOB\_TITLE} entities in free text.

\paragraph{Groups of Job Titles (\textsc{JOB\_TITLE\_GROUP})}~\\
The entity \textsc{JOB\_TITLE\_GROUP} refers to higher-level occupational categories that include multiple related job titles. This entity is intended to take the hierarchical structure of the occupations into consideration.

The German Classification System of Occupations KldB defines a five-level hierarchy of groups, represented by five-digit-codes. The first digit denotes the general occupational area (\emph{Berufsbereich}), down to the fifth digit, which specifies the requirement level or complexity of the occupational activity (\emph{Berufsgattung}).

Each specific job title is allocated to the lowest of these five groups, \eg, the profession of \emph{pharmaceutical technical assistant} (\emph{Pharmazeutisch-technischer Assistent}; code 81822-105) is assigned to the group 81822 \cfcite{KldB2020}:

\textit{\begin{itemize}
    \item[] 8: Health, social services, teaching and education
    \item[] 81: Medical health professions
    \item[] 818: Pharmacy
    \item[] 8182: Professions in pharmaceutical technical assistance
    \item[] 81822: Professions in pharmaceutical technical assistance---specialized activities
\end{itemize}
}\bigskip

The most common job title group in the annotated data is \enquote{Facharbeiterberufe} (\emph{skilled labor occupations}).

\paragraph{Skills (\textsc{SKILL})}~\\
Skills are mental or physical prerequisites for performing an occupation. In the German language, there is a subtle distinction between potential, more general skills (\emph{Fähigkeiten}) and learned skills (\emph{Fertigkeiten}). According to the pedagogical understanding of competence, both types of skills, complemented by knowledge, constitute competences~\cfcite{Hartig2008, Linten2015, Weinert2001}.
The entity \textsc{SKILL} combines both types of skills.

A specific challenge when recognizing skills is the presence of multi-word expressions, either describing a skill in very general terms or representing a conjunction of several skills~\cfcite{Nguyen2024}; \eg, \enquote{Aufmerksamkeit, Geduld und Fingerspitzengefühl} (\emph{attentiveness, patience, and tact}).
The most frequently mentioned skill is \enquote{attentiveness} (\emph{Aufmerksamkeit}).

\paragraph{Training Subjects (\textsc{SUBJECT})}~\\
Training subjects are course topics during vocational education (\emph{Ausbildungsfächer}, \emph{Lehrgänge}) and are typically listed in curricula (\emph{Stundentafeln}).
As entity \textsc{SUBJECT}, specific course titles should be recognized but not detailed lesson contents.
The most frequent subject in the annotated data is \enquote{Werkstoffe} (\emph{work materials}).

\paragraph{Working Activities (\textsc{ACTIVITY})}~\\
The entity \textsc{ACTIVITY} refers to specific work activities of a certain occupation.
Only concrete actions should be annotated as \textsc{ACTIVITY}, not generic phrases.
The most frequently mentioned working activity is \enquote{Reklamationen bearbeiten} (\emph{process complaints}).


\section{\uppercase{Related Work}}
\label{sec:related_work}

In this section, we review existing approaches in the fields of \gls{NER}, \gls{NAT}, and applications to \gls{VET}.


\subsection{Named Entity Recognition (NER)}
\label{sec:related_work_ner}

\gls{NER} aims to identify named entities in text according to predefined categories \citep[cf.][]{Nadeau2007}.
\gls{NER} can be considered a special case of the sequence labeling problem, which generally involves labeling tokens from an input sequence~\citep[cf.][]{TjongKimSang2003}.
A distinction can be made between classical \emph{heuristic-based} and \emph{deep learning-based} \gls{NER} methods.

\paragraph{Heuristic-based Approaches}~\\
Among heuristic-based approaches, a further differentiation exists between \emph{rule-based} and \emph{statistical} methods:
Rule-based methods, including pattern matching, rely on handcrafted rules or regular expressions to identify entities \citep[cf.][]{Eftimov2017}.
These methods require explicit lexicons, \ie, precompiled lists of known entities. 
Statistical methods, such as \glspl{HMM} \citep{Bikel1999} and \glspl{CRF} \citep{McCallum2003}, use sequence labeling techniques based on hand-engineered features \cfcite{Jurafsky2025}.

\glsreset{HMM}
\glsreset{CRF}
\glspl{HMM} model sequences by transition matrices, which represent the joint probability of transitions between hidden input states and observable output states \cfcite{Rabiner1989}.
\glspl{CRF} are undirected graphical models that compute the conditional probability of a sequence of labels given a corresponding input sequence \cfcite{Lafferty2001}.
Algorithms such as the Viterbi algorithm~\citep{Viterbi1967} are commonly used to determine the most likely sequence of states or labels, for both \glspl{HMM} and \glspl{CRF}.

\paragraph{Deep Learning-based Approaches}~\\
In contrast, deep learning-based methods apply modern learning techniques based on Deep Neural Networks (\glsplural{DNN}).
These can be further distinguished based on the model architecture:
\emph{LSTM-based models}, particularly bi-directional \glsplural{LSTM}, are widely used to capture context and are often combined with \glspl{CRF} for sequence decoding \citep[cf.][]{Huang2015, Lample2016}.
These approaches typically require word and character embeddings, such as GloVe, FastText, or FLAIR~\citep[cf.][]{Pennington2014, Bojanowski2017, Grave2018, Akbik2018}. 

\emph{Transformer-based models}, such as BERT and numerous variants, currently represent the state of the art \citep[cf.][]{Devlin2019}.
Transformer architectures consist of stacked self-attention layers that model pairwise dependencies between tokens, enabling each token to access information from the entire sequence while remaining highly parallelizable during training \cfcite{Vaswani2017}. These models do not require explicit dictionaries, but rather rely on contextual understanding learned from large corpora.
Therefore, it is common practice to fine-tune them on domain-specific data, allowing the model to implicitly learn and recognize relevant entities.


\subsection{Noise-aware Training (NAT)}
\label{sec:related_work_nat}

\glsreset{NAT}
\gls{NAT} refers to a set of methods aiming to make downstream \gls{NLP} systems, such as \gls{NER}, robust to noise introduced upstream, \eg, \gls{OCR} errors.
The core idea of \gls{NAT} is to inject realistic errors into the training data so that the model learns representations that are less sensitive to such perturbations \citep[cf.][]{Namysl2020}.
In practice, this involves injecting various \gls{OCR}-specific errors, such as character substitutions, deletions, insertions, and spacing errors.

\citet{Namysl2020} proposed a \gls{NAT} approach for sequence labeling, using artificial perturbations to improve robustness to both \gls{OCR} errors and misspellings on \gls{NER} tasks, without sacrificing performance on clean text.

\citet{Xu2021} injected synthetic \gls{OCR} noise prior to text classification, enabling the model to handle noisy transcripts.

\gls{NAT} is closely related to \emph{Empirical Error Modeling}, which empirically analyzes typical \gls{OCR} errors, in order to subsequently inject realistic perturbations~\cfcite{Namysl2021}.

\emph{Document Noise Modeling} complements noise-aware techniques by explicitly modeling the distortions that commonly arise in document processing scenarios. \emph{TrOCR}~\citep{Li2021} introduces synthetic document degradations (such as blur, low resolution, and scanning artifacts) during pre-training of its encoder–decoder architecture. Similarly, OCR-free models, such as Donut~\citep{Kim2022}, apply controlled perturbations to document images, enabling the model to learn text extraction directly from pixels under noisy conditions.

\emph{Self-Supervised OCR Correction} goes beyond NAT and aims to explicitly \emph{learn to correct} noisy OCR outputs.
Recent methods adopt denoising sequence-to-sequence architectures, often based on pre-trained language models such as BART or T5~\cfcite{Singh2019, Saluja2021, Chen2021}.
Other approaches leverage confidence-aware objectives~\cfcite{Boubt2022} or masked language modeling to refine noisy OCR text~\cfcite{Huang2022}.


\subsection{Application to VET}
\label{sec:related_work_vet}

A limited number of publications have addressed \gls{NER} in labor market data, and in VET documents in particular.
Most publications refer to the recognition of either \emph{job titles} or \emph{skills}.

\paragraph{Job Titles}~\\
\citet{Reiser2025} detected and classified occupations in German texts, comparing a rule-based approach with a language model-based approach.

\citet{Safikhani2023} fine-tuned BERT and \mbox{GPT-3} for the automated extraction of German occupations.

\citet{Decorte2021} proposed JobBERT, a deep learning-based model for recognizing job titles and linking them to ESCO classes. Starting from BERT as a pre-trained model, the recognition is based on skill information, showing that skill descriptions are an essential component for job title recognition.

It should be noted that the objective of this paper differs from \emph{occupation coding}, which is a text classification problem for matching a given job title against a pre-defined list, such as the KldB \cfcite{Gweon2017, Schierholz2014, Schierholz2021}. In contrast, the goal of this work is to identify job titles (and other entity types) in free text.

\paragraph{Skills}~\\
\citet{Wang2025b} presented a pipeline for skill extraction from job advertisements.

\citet{Li2023b} and \citet{Nguyen2024} recently employed \glspl{LLM} for skill extraction.

\citet{Tabares2018} provided methods for constructing competency-based ontologies.

\citet{Huang2020} proposed \emph{Skill2Vec}, an approach for numerically embedding skills beyond ESCO-specific modeling.

Skill recognition is closely related to \emph{competency modeling}, which has a long tradition in education research. Related work includes foundational frameworks such as O*NET~(\citeyear{ONET2020}), the HR-XML competency schema~(\citeyear{HRXML2007}), and the IEEE Reusable Competency Definitions standard~(\citeyear{IEEE2008}).
Regarding German competency models, \citet{Tiemann2024b, Weinert2001, Hartig2008, Linten2015} provide an overview. 

\paragraph*{}
Most publications use \gls{NER} to recognize a single type of entity, such as job titles or skills. They do not take into account the semantic connection between different entity types. In contrast, in this work multiple entity types are deliberately recognized. Searching for explicit expressions, such as job titles, is error-prone; incorporating additional contextual and implicit information, as our long-term goal, aims to improve the information extraction of the pipeline to be built.


\section{\uppercase{Proposed Method}}
\label{sec:method}

OCR noise can significantly affect downstream tasks like \gls{NER}.
Therefore, we propose a \gls{NAT} approach to increase robustness of \gls{NER} against \gls{OCR} errors.

\citet{Namysl2023a} studied the robustness of document processing pipelines in depth.
\emph{Domain-specific retraining}, \ie, fine-tuning an existing model on a set of domain-specific documents, often already yields significant improvements.
A more advanced approach is~\emph{\gls{NAT}}.

An additional challenge arises from the variability of entities. One reason is the temporal evolution of entities. The historical \gls{VET} documents originate from different decades and from two German states with different socio-political systems. Terms (such as job titles) and formulations (such as activity descriptions) may change over time.
Further variations arise from language-specific characteristics. \citet{Reiser2025} discusses the unique challenges of the German language, using occupational titles as an example. A typical source of variation are gender-specific forms of job titles.

Due to all these challenges, rule-based approaches are not promising.
Instead, three transformer-based models are trained, building on BERT as a pre-trained state-of-the-art model, specifically adapted for recognizing entities in German \gls{VET} documents and capturing the domain's vocabulary and characteristics.

\subsection{Proposed Models}
\label{sec:models}
The proposed method involves training three models: \emph{noisy}, \emph{clean}, and \emph{artificial}.
The \emph{noisy} model is based on training data with uncorrected \gls{OCR} errors.
For the \emph{clean} model, \gls{OCR} errors have been corrected prior to training.
The \emph{artificial} model builds upon the \emph{clean} model; applying \gls{NAT} techniques, realistic \gls{OCR} errors (insertions, deletions, substitutions) are injected synthetically.

Please note that this results in different amounts of training data.
The \emph{noisy} and \emph{clean} models both use the same original amount of training data.
For the \emph{artificial} model, in each correctly spelled word one typical error -- a \emph{substitution}, \emph{deletion}, or \emph{insertion} -- is injected. This results in doubling the amount of data. \Cref{tab:training_data_models} lists the amount of training data used for each of the three models. 

For fair comparability, in this experiment, the amount of training data was increased only moderately. In practical applications, however, \gls{NAT} provides a simple way of multiplying the dataset size as a form of \emph{data augmentation}.

\begin{table}[t]
    \caption{Amount of training data by model.}
    \label{tab:training_data_models}
    \centering
    \begin{tabularx}{0.7\textwidth}{Xrr}
        \toprule

        \textbf{Model} & \textbf{Entities} & \textbf{Tokens}\\

        \begin{filecontents*}{data/data_amount.csv}
model;entities;tokens
Noisy;1\,058;7\,280
Clean;1\,058;7\,280
Artificial;2\,116;14\,560
\end{filecontents*}
\csvreader[
        head to column names,
        separator=semicolon,
        late after line=\\
        ]{data/data_amount.csv}
        {}
        {\emph{\model} & \entities & \tokens}

        \bottomrule
    \end{tabularx}
\end{table}

\paragraph{Transfer Learning}~\\
During model training, \emph{transfer learning} is applied. Transfer learning is a common concept in which pre-trained models, which have been trained on large datasets, are fine-tuned for target tasks on smaller, domain-specific datasets.
\citet{Namysl2019} demonstrated that a backbone model (in their case, for \gls{OCR}) can be significantly improved by adding only a small amount of data.

\pagebreak[3]

\subsection{Training Data}
\label{sec:training_data}

\begin{table}[t]
    \caption{Number of annotated entities.}
    \label{tab:annotated_entities}
    \centering
    \begin{tabularx}{0.7\textwidth}{Xr}
        \toprule
        \textbf{Entities} & \textbf{1509}\\
        \hspace{1em}\textsc{JOB\_TITLE} & 916\\
        \hspace{1em}\textsc{JOB\_TITLE\_GROUP} & 121\\
        \hspace{1em}\textsc{SKILL} & 54\\
        \hspace{1em}\textsc{SUBJECT} & 167\\
        \hspace{1em}\textsc{ACTIVITY} & 251\\
        \bottomrule
    \end{tabularx}
\end{table}

The training data consists of historical VET documents from East and West Germany as well as reunified Germany,
especially training regulations (\emph{Ausbildungs-/ Berufsordnungen}) and systematic listings of job titles (\emph{Systematische Gliederungen}),
covering the period since 1976,
with 68 document pages in total, annotated for both OCR errors and entity recognition.
\Cref{tab:annotated_entities} lists how many occurrences of each entity are present in the annotated data.

The ideal training data would contain every possible job title, job title group, skill, subject, and activity, ideally multiple times, with and without OCR errors, requiring millions of observations. As this is unattainable, the goal is to provide a sufficiently large and realistic dataset so that the model can generalize from these data for future predictions.

\paragraph{Additional Training Data}~\\
To increase the accuracy, additional training data has been integrated:

For \emph{job titles}, lists of search terms provided by the Federal Employment Agency\footnote{See\,{}\url{https://www.arbeitsagentur.de/institutionen/dkz-downloadportal}.}, linking free text occupational titles to the KldB, was used  as additional data.

For \emph{job title groups}, the KldB \citep{KldB2020} with its hierarchical structure, containing numerous umbrella terms for occupations, was used.

For \emph{skills}, the ESCO taxonomy\footnote{See\,{}\url{https://esco.ec.europa.eu/en/use-esco/download}; version 1.2.0.} \citep{ESCO}, containing more than 30\,000 standardized skills, has been used.

However, it is important to note that all this additional training data differs from real-world data in two important respects:
Every word belongs to an entity; the data contains no \enquote{O} labels stating that the corresponding token is not part of any entity. Besides, the data contains no \gls{OCR} noise.

Therefore, the additional training data cannot be used directly for the final models. Instead, multi-stage fine-tuning is applied, introducing an intermediate pre-training stage trained on this data, as described in \Cref{sec:multi_stage}.
In a similar manner, \citet{Schierholz2021} improved the performance of algorithms (in their case, for occupation coding) by adding data from indices. 

\subsection{Error Injection}
\label{sec:error_injection}

\begin{table}[t]
    \caption{Most frequent OCR errors.}
    \label{tab:ocr_errors}
    \centering
    \begin{tabularx}{0.5\textwidth}{@{}X@{\hspace{0.3em}}X@{}X@{\hspace{0.3em}}r@{}}
        \toprule
        \textbf{Recognized} & \textbf{Correct} & \textbf{Type} & \textbf{Frequency}\\

        \begin{filecontents*}{data/ocr_error_analysis.csv}
gt;ocr;type;cnt
-;;deletion;311
,;;deletion;240
e;;deletion;106
.;;deletion;104
n;;deletion;77
r;;deletion;60
t;;deletion;49
s;;deletion;48
u;;deletion;47
i;;deletion;40
m;n;substitution;39
a;;deletion;36
a;ä;substitution;33
l;i;substitution;32
\end{filecontents*}
\csvreader[
        head to column names,
        separator=semicolon,
        late after line=\\
        ]{data/ocr_error_analysis.csv}
        {}
        {\ocr & \gt & \type & \cnt}

        \bottomrule
    \end{tabularx}
\end{table}  

Applying \gls{NAT}, one error, a \emph{substitution}, \emph{deletion}, or \emph{insertion}, is injected into each correctly spelled word.

When injecting these artificial errors, typical mistakes made by \gls{OCR} engines are explicitly taken into account; such errors are inserted with a higher probability. For this purpose, an empirical error analysis was conducted.

\paragraph{Empirical OCR Error Analysis}~\\
A qualitative error analysis shows that, in the historical \gls{VET} documents, the following types of \gls{OCR} errors occur particularly frequently:
(i) punctuation not recognized (most common error);
(ii) individual letters misrecognized (\eg, \enquote{l} recognized as \enquote{i});
(iii) difficulties in recognizing digits (\eg, \enquote{0} recognized as \enquote{o});
(iv) scanning artifacts, \ie, spurious single characters (\eg, stray commas or apostrophes);
(v) confusions between periods and commas (likely due to typewritten fonts and document aging).
\Cref{tab:ocr_errors} lists the 15 most frequent \gls{OCR} errors observed in the annotated historical \gls{VET} documents.  


\subsection{Multi-stage Fine-tuning}
\label{sec:multi_stage}

To integrate the additional training data described in \Cref{sec:training_data}, we introduce an intermediate pre-training stage, applying \emph{\gls{DAPT}}~\cfcite{Gururangan2020}.

\pagebreak[1]

\begin{figure}[!ht]
    \centering
    \includegraphics[width=0.6\textwidth]{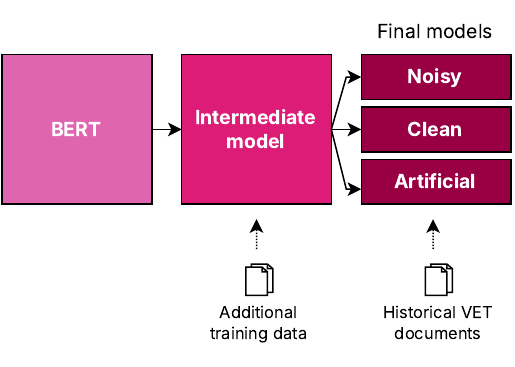}
    \caption{Multi-stage training process.}
    \label{fig:multi_stage}
    \vspace{1em}
\end{figure}

\pagebreak[1]

\begin{samepage}
\Cref{fig:multi_stage} shows the multi-stage training process: as base model, we use a German, cased BERT variant, \texttt{dbmdz/bert-base-german-cased}\footnote{See\,{}\url{https://huggingface.co/dbmdz/bert-base-german-cased}.}, released by the Bavarian State Library, with weights initialized from the original BERT values.
\end{samepage}

The intermediate model is then trained on the large amount of additional training data. This constitutes the first step of adapting the model to the \gls{VET} domain, applying \gls{DAPT}, in order to specifically recognize the entities mentioned in \Cref{sec:background-entities}. Finally, the three model variants \emph{noisy}, \emph{clean}, and \emph{artificial} are trained on the annotated real-world data.

\citet{Hendrycks2019} showed that fine-tuning models in this way can significantly increase robustness, which is particularly important for this use case due to \gls{OCR} errors.


\subsection{Implementation Details}
\label{sec:method_implementation_details}

The framework is built on PyTorch \citep{Paszke2019} and the Hugging Face Transformers library \citep{Transformers}.

The computing costs were negligible, as we relied on a pre-trained BERT model; the training could be completed on a local workstation equipped with an NVIDIA GPU within short time.

The annotated dataset is split into training, test, and validation sets (70{:}20{:}10), all having the same distribution of entities.
Splitting is performed at segment level, not at token level, to avoid breaking multi-word entities. 

\paragraph{Bias}~\\
When using pre-trained models, it is important to note that these may be subject to bias present in their training data and affecting their predictions. This bias is carried over to the fine-tuned model. The additional training data may also be subject to bias~\mbox{\cfcite{Wang2023}}.

\paragraph{Hyperparameters}~\\
When training the final three models, we search for the lowest validation loss using the following hyperparameters:
\emph{Learning rate scheduling} is particularly relevant for Transformer-based models, which are often sensitive to large parameter shifts. During a warm-up phase (10\,\% of the steps), the learning rate is gradually increased up to the target rate of $2\cdot10^{-5}$.
In the subsequent decay phase, the learning rate is decreased to a minimum value as training progresses (following an inverse root function). Especially when fine-tuning pre-trained models, smaller learning rates allow for more refined adjustments.

The maximum \emph{number of epochs} is 25. However, analysis (see \Cref{sec:precision_recall}) shows that this limit is never reached: The loss function begins to converge after approximately four epochs (see also \Cref{fig:precision_recall}); early stopping is applied when no progress is observed for five consecutive epochs.

The best-performing model over all epochs, measured by the $F_1$ score (micro-averaged on entity level), is saved.

\paragraph{Class Weighting and Oversampling}~\\
To compensate for the significant imbalance, we apply \emph{class weighting} and, additionally, \emph{oversampling}. The majority of labels is \enquote{O}, stating that the corresponding token is not part of any entity. The loss function would be dominated by these labels, and the model could degenerate into predicting only \enquote{O} (class collapse).
For class weighting, we use balanced (inverse-frequency) weighting, then normalize the weights, and clamp them to avoid extreme values.
Oversampling, during training, replicates token windows that contain positive labels (non-\enquote{O}) and thereby increases the relative frequency of positive examples presented to the model. This artificially rebalances the class distribution and strengthens the loss signal for rare entities.

For pre-training the intermediate model, slightly different hyperparameters and settings have been used: The additional training data for the intermediate model contains no \enquote{O} labels. Therefore, no oversampling, weight balancing, and clamping has been applied.
As the dataset for the intermediate model is quite large (containing 557\,749 token-label pairs), the validation set has been limited to 10\,000 tokens.

\paragraph{Annotations: BIO Format}~\\
For annotating entities, the \emph{BIO} format is used, in which each word (token) is assigned a tag. The tag \enquote{B} indicates that the token marks the beginning of an entity span; \enquote{I} indicates that the token lies inside an entity span; \enquote{O}, outside any entity span.


\paragraph{Token Simplification}~\\
Regarding token simplification, two contradictory positions exist.
Some authors recommend normalizing tokens (lowercasing, stemming, removing special characters, and stopwords) before \gls{NER} to reduce noise~\cfcite{Gweon2017, Baldwin2015}.
In contrast, other authors argue that heavy preprocessing can harm \gls{NER}, as it removes strong cues~\cfcite{Mayhew2020, Alzahrani2021}. For instance, uppercase can be an informative signal for names and organizations; morphological endings (in German, \eg, \emph{-in}, \emph{-meister}, and \emph{-en} in plural) help to predict entity boundaries; stemming would collapse endings.
We did not apply token simplification because experiments showed that, in our use case, it is not beneficial (\FOne score decreasing from 77.5\,\% to 63.3\,\%).

\pagebreak[3]


\section{\uppercase{Evaluation}}
\label{sec:evaluation}
In this section, we present the results of our evaluation, including a comparison of the three final models and their performance across different entity types.

A direct comparison with state-of-the-art \gls{NER} models, such as BERT, is not possible, as these are trained for different entity types (persons, locations, etc.), not for \gls{VET} entities.

Likewise, a comparison with other methods on \gls{VET} data is difficult.
Some of the authors mentioned in \Cref{sec:related_work_vet} published recognition rates for the recognition of occupational titles; however, on different datasets and for the task of matching job titles against classification systems, not for identifying job titles in continuous text. Therefore, these recognition rates are not comparable. Thus, we report performance metrics as context, not for establishing a benchmark.


Common evaluation metrics for \gls{NER} are \emph{precision}, \emph{recall}, and the \emph{\FOne score}. These are based on the number of \emph{true positives} $\TP$, \emph{false positives} $\FP$, and \emph{false negatives} $\FN$:  

\begin{samepage}
\begin{equation*}
    \begin{aligned}
        \Pr & = \frac{\TP}{\TP + \FP} \\[0.5em]
        \Re & = \frac{\TP}{\TP + \FN} \\[0.5em]
        F_1 & = \frac{2 \cdot \Pr \cdot \Re}{\Pr+ \Re}
    \end{aligned}
\end{equation*}
\end{samepage}


\subsection{Precision and Recall over Time}
\label{sec:precision_recall}

\begin{figure}[!ht]
    \centering
    \includegraphics[width=0.7\textwidth]{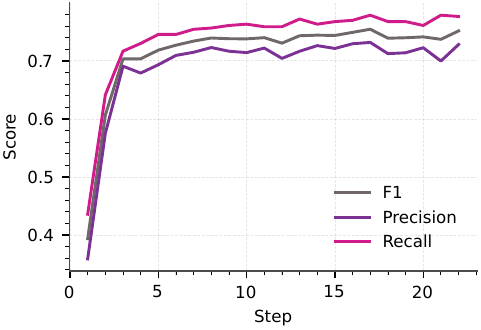}
    \caption{Precision, recall, and \FOne score over time.}
    \label{fig:precision_recall}
    \vspace{1em}
\end{figure}

\Cref{fig:precision_recall} shows precision, recall, and \FOne score of the \emph{noisy} model during training.
Throughout all epochs, precision and recall remain close to each other, without notable deviations. Recall is continuously slightly higher than precision.

From a theoretical perspective, a higher recall indicates that the model is able to locate entities within a sentence but fails to capture the exact span boundaries.


\subsection{Model Comparison}
\label{sec:evaluation-comparison_models}

To ensure comparability, all three models were trained with the same parameters and on the same training, test, and validation set. The training data differs only in its characteristics: using uncorrected \gls{OCR} results (\emph{noisy} model), using manually corrected \gls{OCR} results (\emph{clean} model), or having synthetic \gls{OCR}-like errors injected (\emph{artificial} model). 

\begin{samepage}
\Cref{tab:accuracy} reports the accuracy for each model. The \emph{artificial} model achieves the highest \FOne score~(77.9\,\%).
\end{samepage}

\pagebreak[2]

Please note that \gls{OCR} correction and synthetically injected noise affect only the training and validation dataset, not the test dataset. Later, during inference, the model will be invoked with noisy inputs; thus, the test data should reflect real-world performance.

Therefore, as expected, the $F_1$ score of the \emph{clean} model is lower than that of the \emph{noisy} model, since the \emph{clean} model is not trained to handle noisy data. This gap captures the impact of noise and the additional robustness provided by the \emph{noisy} model.

It is noteworthy that the \emph{noisy} and the \emph{clean} model have comparable precision, but the latter has a lower recall.
This is presumably caused by data shift: The \emph{clean} model has been trained on clean tokens but evaluated on noisy text. When first being confronted with perturbed tokens at test time, the model becomes conservative and skips boundary spans, leading to more false negatives and, thus, a lower recall.

The \emph{artificial} model (this is a strong result) outperforms the \emph{noisy} model, showing that synthetic data injection can even improve the original accuracy.

As described in \Cref{sec:models}, for this experimental evaluation, the training set size was only moderately increased by noise injection, in order to maintain comparability. For practical application, the dataset could be substantially enlarged via \gls{NAT}, presumably further improving the \emph{artificial} model.

The described basic order---the \emph{artificial} model achieving the best results, followed by the \emph{noisy} and the \emph{clean} model---remained consistent across multiple repetitions of the evaluation with different random seeds for dataset splitting and model initialization.

\begin{table}[t]
    \caption{Accuracy per model.}
    \label{tab:accuracy}
    \centering
    \begin{tabularx}{0.7\textwidth}{Xrrr}
        \toprule
        \textbf{Model} & \textbf{Precision} & \textbf{Recall} & \textbf{\FOne score}\\

        \begin{filecontents*}{data/eval_accuracy.csv}
model;pre;rec;fone
Noisy;73.4\,\%;82.0\,\%;77.5\,\%
Clean;73.5\,\%;76.9\,\%;75.2\,\%
Artificial;75.7\,\%;80.3\,\%;77.9\,\%
\end{filecontents*}
\csvreader[
        head to column names,
        separator=semicolon,
        late after line=\\
        ]{data/eval_accuracy.csv}
        {}
        {\emph{\model} & \pre & \rec & \fone}

        \bottomrule
    \end{tabularx}
\end{table}


\subsection{Accuracy per Entity}
\label{sec:evaluation-comparison_entities}

\Cref{tab:accuracy_entities} lists the \FOne score per entity using the \emph{artificial} model. The different numbers of annotated entities, as shown in \Cref{tab:annotated_entities}, lead to different recognition rates across entities. Additionally, some entities are inherently more complex to detect. \emph{Job titles} can be recognized with the highest accuracy.

\begin{table}[t]
    \caption{Accuracy per entity.}
    \label{tab:accuracy_entities}
    \centering
    \begin{tabularx}{0.7\textwidth}{Xr}
        \toprule
        \textbf{Entity} & \textbf{\FOne score}\\

        \begin{filecontents*}{data/accuracy_entities.csv}
entity;fone
JOB\_TITLE;87.9\,\%
JOB\_TITLE\_GROUP;76.5\,\%
SKILL;38.5\,\%
SUBJECT;58.8\,\%
ACTIVITY;40.0\,\%
\end{filecontents*}
\csvreader[
        head to column names,
        separator=semicolon,
        late after line=\\
        ]{data/accuracy_entities.csv}
        {}
        {\textsc{\entity} & \fone}

        \bottomrule
    \end{tabularx}
\end{table}

\paragraph{Skills}~\\
For \emph{skills}, there exists a second challenge, besides the small number of annotated entities: the skills in the annotated \gls{VET} documents differ, in terms of their content, from those in the additional training data. The training regulations in the \gls{VET} dataset contain \emph{prerequisite} skills for practicing an occupation, both physical and mental ones; \eg, \enquote{Fingerfertigkeit} (\emph{dexterity}). In contrast, the ESCO data used as additional training data predominantly contains \emph{learnable} skills to be learned during training (\eg, programming languages). As a result, the recognition rate for \textsc{SKILL} is comparatively low.

This highlights a central challenge of multi-stage fine-tuning when the annotation conventions for an entity differ across datasets. In future work, instead of a single overarching \textsc{SKILL} entity, we aim to distinguish more finely between prerequisite and learnable skills.

\paragraph{Subjects and Activities}~\\
For \emph{subjects} and \emph{activities}, no additional training data (and, thus, no pre-trained weights from the intermediate model) existed. Consequently, the recognition rates for both entities are lower. The influence of multi-stage fine-tuning becomes clear when comparing the \FOne scores of \emph{subjects} and \emph{activities} to those of \emph{job titles}, which benefit from a large amount of additional training data derived from the KldB.


\subsection{Confusion of Entities}
\label{sec:evaluation-confusion_entities}

\begin{figure}[!ht]
    \centering
    \vspace{1em}
    \includegraphics[width=0.7\textwidth]{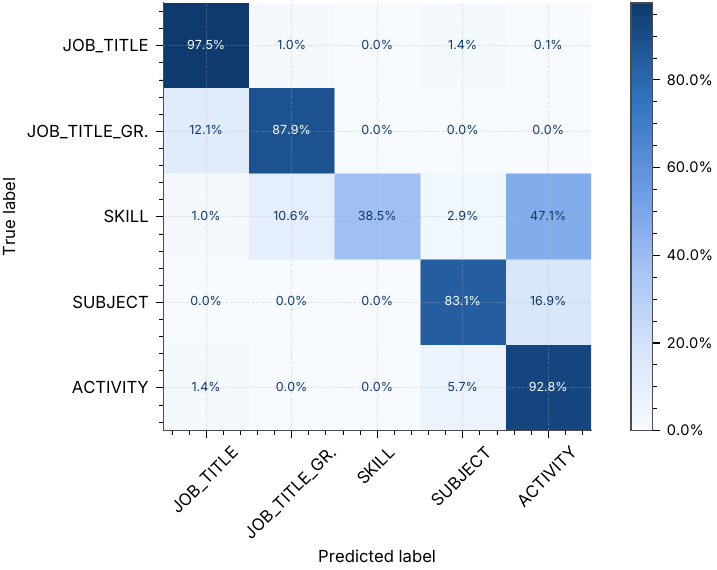}
    \caption{Confusion matrix for entity recognition.}
    \label{fig:confusion_matrix}
    \vspace{1em}
\end{figure}

\Cref{fig:confusion_matrix} shows the confusion matrix for entity recognition (token-level confusions; without \enquote{O} labels) using the \emph{noisy} model.

The confusion matrix indicates that there are particular difficulties in distinguishing between specific entities:

For \emph{job titles} versus \emph{job title groups}, it is not always clear whether a term refers to a specific occupation or to a higher-level category; \eg, \enquote{Facharbeiter} (\emph{skilled worker}).

For \emph{skills} versus \emph{activities}, the challenge arises because the ESCO, as described in \Cref{sec:evaluation-comparison_entities}, predominantly contains \emph{learnable} skills, which are similar to working activities.

\pagebreak[2]

For \emph{activities} versus \emph{subjects}, there is an elevated risk of confusion because subjects are an umbrella category for course topics that may contain multiple activities.

\pagebreak[3]


\section{\uppercase{Conclusion}}
\label{sec:conclusion}

This publication investigated the complex task of recognizing named entities in historical German \gls{VET} documents, highlighting contextual challenges.
A particular focus has been put on \gls{OCR} errors that propagate to downstream tasks, such as \gls{NER}. We introduced a robust, noise-aware \gls{NER} approach tailored to \gls{VET} documents and demonstrated improved performance compared with baseline and domain-specific models.
Three models have been trained and compared: \emph{noisy}, \emph{clean}, and \emph{artificial}, capturing uncorrected \gls{OCR} output, corrected \gls{OCR} output, and synthetically perturbed text. The \emph{artificial} model achieved the best results ($F_1$ score: 77.9\,\%).

We further compared the accuracy across different entity types. For \emph{job titles}, the highest recognition rate was achieved ($F_1$ score: 87.9\,\%). In a qualitative analysis, we also examined confusions and intrinsic difficulties in recognizing different entity types.

Overall, the results support the effectiveness of noise-aware training for historical \gls{VET} documents.

\pagebreak[3]


\subsection{Limitations and Future Work}
\label{sec:conclusion-limitations-future}

The results motivate continued work on developing a complex document processing pipeline for VET documents, especially integrating further preprocessing methods, layout analysis and table recognition, to improve recognition quality.

\paragraph{OCR Errors}~\\
\gls{OCR} engines, such as Tesseract, usually come with built-in preprocessing methods, \eg, for binarization and skew correction.

The historical \gls{VET} documents held by the BIBB have been professionally digitized, so varying illumination during scanning is not an issue. However, owing to their age, many documents exhibit yellowed paper and fading text. Such artifacts can cause standard binarization methods to fail; more advanced, locally adaptive binarization methods are necessary.

The same holds for skew correction: standard methods may fail in case of noisy backgrounds, low contrast, or large rotation angles.

Therefore, one future step is the implementation of custom preprocessing methods prior to \gls{OCR}, as well as post-correction strategies to improve OCR output quality. \glspl{VLM}, such as LayoutLMv3 \citep{Huang2022}, are promising for incorporating layout-based features as well.

For this study, Tesseract as a state-of-the-art OCR engine has been used; an open avenue for future work is to evaluate the stability of our findings across different OCR systems.

\paragraph{Complex Page Layouts}~\\
Another key challenge for \gls{NER} are complex page layouts. In some cases, entities are not recognized due to line breaks, when the reading order could not be determined correctly. Therefore, an essential next goal is to integrate layout analysis and table recognition into the document processing pipeline to handle complex page layouts prior to \gls{NER}.

\paragraph*{}
The automated extraction of occupational information from documents remains an important concern. In this publication, foundations have been created for an expandable pipeline for the analysis of \gls{VET} documents.

\bigskip

\begin{samepage}
{\footnotesize
    \noindent\textbf{Author Contributions:}
    Conceptualization: A.E.;
    Data Curation: A.E.;
    Methodology: A.E.;
    Writing, Review, and Editing: A.E., J.D.;
    Investigation: J.D.;
    Visualization: A.E.;
    Supervision: J.D.\\~\\
    All hyperlinks were last verified on 19/09/2025.
}
\end{samepage}

\pagebreak

\bibliographystyle{apalike}
\bibliography{references}

\end{document}